\documentclass{llncs}
\usepackage{makeidx}  
\usepackage[pdftex]{graphicx}
\usepackage{latexsym}
\usepackage{verbatim}
\usepackage{tabularx}
\usepackage{wrapfig}
\usepackage{amsmath}
\usepackage{multirow}

\newcommand{\shortcite}[1]{\cite{#1}}

\begin{document}
\frontmatter          
\pagestyle{headings}  
\mainmatter              
\title{Generating Different Story Tellings from Semantic Representations of Narrative}
\titlerunning{Generating Different Story Tellings}  

\author{Elena Rishes\inst{1} \and Stephanie Lukin\inst{1} \and David K. Elson\inst{2} \and Marilyn A. Walker\inst{1}}

\authorrunning{Rishes, Lukin, Elson and Walker} 
\institute{Natural Language and Dialogue Systems Lab \\ University of California Santa Cruz, Santa Cruz\\
\email{\{erishes, slukin, maw\}@soe.ucsc.edu} \\
\and
Columbia University, New York City \\
\email{delson@cs.columbia.edu}
}

\maketitle 

\begin{abstract}
In order to tell stories in different voices for different audiences,
interactive story systems require: (1) a semantic representation of story
structure, and (2) the ability to automatically generate story and
dialogue from this semantic representation using some form of Natural
Language Generation ({\sc nlg}).  However, there has been limited
research on methods for linking story structures to narrative
descriptions of scenes and story events.  In this paper we present an
automatic method for converting from Scheherazade's {\sc story
  intention graph}, a semantic representation, to the input required by the {\sc personage} {\sc
  nlg} engine. Using 36 Aesop Fables distributed in DramaBank, a collection of story encodings, we train translation rules on one story and then test
these rules by generating text for the remaining 35.  The results are measured
in terms of the string similarity metrics Levenshtein Distance and BLEU score. The results show that we can generate
the 35 stories with correct content: the test set stories on average
are close to the output of the Scheherazade realizer, which
was customized to this semantic representation. We provide some examples of story variations generated by {\sc personage}. In future work, we will experiment with measuring the quality of the same stories generated in
different voices, and with techniques for making storytelling
interactive.  \keywords{Semantic Narrative Representation, Natural
  Language Generation, StoryVariation }
\end{abstract}

\section{Introduction}

Sharing our experiences by storytelling is a fundamental and prevalent
aspect of human social behavior.  A critical aspect of storytelling
``in the wild" is that it is socially interactive and situation
dependent.  Storytellers dynamically adjust their narratives to the
context and their audience, telling and retelling the same story in
many different ways depending on who the listener is. For example,
storytellers tell richer stories to highly interactive and responsive
addressees \cite{Thorne87}, and
stories told by young adults ``play" to the audience, repeatedly
telling a story until a satisfactory peer response is found
\cite{ThorneMcLean03}.

\begin{figure}[t!hb] 
\begin{center}
\begin{scriptsize}
\begin{tabular}{|p{1.5in}|p{1.6in}|p{1.5in}|}
\hline
{\bfseries Original Story}  & {\bfseries Scheherazade} & {\bf Personage} \\ \hline \hline
A hungry Fox saw some fine bunches of Grapes hanging from a vine that
was trained along a high trellis, 
and did his best to reach them by
jumping as high as he could into the air. 
But it was all in vain, for
they were just out of reach: so he gave up trying, and walked away
with an air of dignity and unconcern, 
remarking, ``I thought those
Grapes were ripe, but I see now they are quite sour."

& Once, a
group of grapes was hanging on a vine and the vine was hanging on a
high trellis.
A hungry fox saw the group of grapes. 
The fox jumped in order to obtain the group of grapes.  
The fox didn't obtain the group of grapes because he wasn't able to reach the group of grapes. 
The fox walked away from the group of grapes with dignity and with unconcern.  
The fox said that he had earlier thought that the group of grapes was ripe and said that he now saw it as being sour.   

&  The group of grapes hung on the vine.
       The vine hung on the trellis.
       The fox saw the group of grapes.
       The fox jumped in order for the fox to obtain the group of grapes.
       The fox did not obtain the group of grapes because the fox was not able to reach the group of grapes.
       The fox walked away from the group of grapes with dignity and unconcern.
       The fox said the fox earlier thought the group of grapes was ripe.
       The fox said the fox now saw the group of grapes was sour. 
\\ \hline
\end{tabular}
\end{scriptsize}
\vspace{-0.1in}
\caption{\label{fox-grapes-example} 
``The Fox and the Grapes" Aesop fable with Generated Versions}
\vspace{-0.4in}
\end{center}
\end{figure}

To have this human capability of telling
stories in different voices for different audiences, interactive story
systems require: (1) a semantic representation of story structure, and (2)
the ability to automatically generate story and dialogue from this
semantic representation using some form of Natural Language
Generation ({\sc nlg}).  However to date, much of the research on
interactive stories has focused on providing authoring tools based
either on simple story trees or on underlying plan-based
representations. In most cases these representations bottom out in
hand-crafted descriptive text or hand-crafted dialogue, rather than
connecting to an {\sc nlg} engine.  

Prior research on {\sc nlg} for story generation has primarily focused
on using planning mechanisms in order to automatically generate story
event structure, with limited work on the problems involved with
automatically mapping the semantic representations of a story and its
event and dialogue structure to the syntactic structures that allow
the story to be told in natural language
\cite{CallawayLester02,Turner94,RiedlYoung94}.  Recent research
focuses on generating story dialogue on a turn by turn basis and
scaling up text planners to produce larger text prose
\cite{CavazzaCharles05,RoweHaLester08,LinWalker11,Walkeretal11}, but
has not addressed the problem of bridging between the semantic
representation of story structure and the {\sc nlg} engine
\cite{RiedlYoung94,RoweHaLester08,CavazzaCharles05,Montfort07,McIntyreLapata09}. An
example of this work is the {\sc storybook} system
\cite{CallawayLester02} which explicitly focused on the ability to
generate many versions of a single story, much in a spirit of our own
work.  The {\sc storybook} system could generate multiple versions of the
``Little Red Riding Hood'' fairy tale, showing both syntactic and lexical
variation. However, {\sc storybook} is based on highly handcrafted
mappings from plans to the {\sc fuf-surge} realizer \cite{Elhadad92}
and is thus applicable only to the ``Little Red Riding Hood'' domain.

To our knowledge, the only work that begins to address the
semantic-to-syntactic mapping within the storytelling domain is Elson's
Scheherazade story annotation tool
\cite{ElsonMcKeown09}. Scheherazade allows na\"{i}ve users
to annotate stories with a rich symbolic representation called 
a {\sc story intention graph}. This representation is robust and
linguistically grounded which makes it a good candidate for a content
representation in an {\sc nlg} pipeline.

In this paper, we present a working model of reproducing different
tellings of a story from its symbolic representation.  In
Sec.~\ref{back-sec} we explain how we build on the {\sc story
  intention graph} representation provided by Scheherazade
\cite{Elson12} and the previous work on {\sc nlg} for interactive
stories based on extensions to the {\sc personage} {\sc nlg} engine
\cite{MairesseWalker11,Walkeretal11}. Sec.~\ref{method-sec} presents
an automatic method for converting from Scheherazade's {\sc story
  intention graph} output to the input required by {\sc
  personage}. Using the corpus of semantic {\sc sig} representations
for 36 Aesop Fables that are distributed in DramaBank \cite{Elson12},
we train translation rules on one story and then test these rules by
generating 35 other stories in the
collection. Figs.~\ref{fox-grapes-example} and~\ref{lion-boar-example}
show two Aesop Fables, with both Scheherazade and {\sc personage}
generated versions. ``The Fox and the Grapes" was the development
story for our work, while ``The Lion and the Boar" was part of the
test set. Fig.~\ref{fox-grapes-retellings} gives a feel for the retellings that {\sc
  personage} 
is capable to produce once it is coupled with Scheherazade's story representation. 
Sec.~\ref{results-sec} demonstrates our evaluation of the 35
generated stories using the measures of Levenshtein Distance and BLEU
score.  Sec.~\ref{conc-sec} summarizes and discusses our future work,
where we aim to experiment with models for narrator's and characters' voices, measures of retellings' quality, and with techniques for making story telling interactive.


\section{Background}
\label{back-sec}

Our work is based on a simple observation: a novel capability for
interactive stories can be developed by bridging two off-the-shelf
linguistic tools, Scheherazade and {\sc personage}. We integrated these tools
in a standard {\sc nlg} pipeline: 
\begin{itemize}
\item Content planner that introduces characters and events
\item Sentence planner that creates linguistic representations for those events
\item Surface realizer that produces text string out of linguistic structures 
\end{itemize}
Scheherazade produces the output we would normally get from a content planner and
{\sc personage} plays the role of the sentence planner
and surface realizer. We developed an
algorithm that creates the semantic linguistic representation from a
conceptual narrative structure provided by Scheherazade and generates
from it using {\sc personage}. Our algorithm acts as an
intermediary producing a semantic-into-syntactic mapping. The
immediate goal for this study was to regenerate directly from stories
annotated with Scheherazade. Our long term goal is to build on the
created infrastructure and turn the intermediary into a conventional
sentence planner, capable of one-to-many semantic-to-syntactic
mappings, i.e. retelling a story in different voices. 

We believe that the combination of Scheherazade and {\sc personage} is
a perfect fit for our long-term goal due to the following aspects of
the two systems. First, Scheherazade representations are already
lexically anchored into WordNet and VerbNet ontologies which allows
for lexical variation. Second, {\sc personage} provides 67 parameters
that make it already capable of generating many pragmatic and
stylistic variations of a single utterance. Below 
 we discuss the functionality of the two systems in more
detail.

\begin{figure}[!h]
\begin{center}
\includegraphics[width=0.9\textwidth]{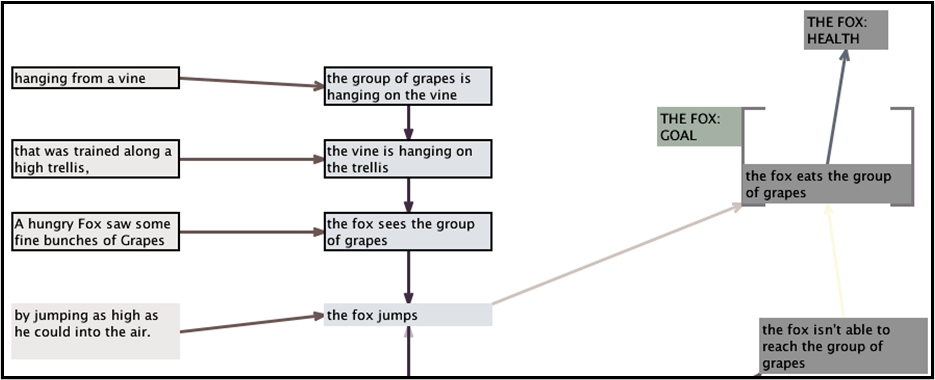}
\vspace{-0.15in}
\caption{\label{foxgrapes-sig} Part of the {\sc Story Intention Graph} ({\sc sig}) for ``The Fox and the Grapes"}
\vspace{-0.3in}
\end{center}
\end{figure}

\noindent{\bf Scheherazade.} Scheherazade is an annotation tool that
facilitates the creation of a rich symbolic representation for narrative texts, using a schemata known as the
{\sc story intention graph} or {\sc sig}
\cite{ElsonMcKeown09}.  An example {\sc sig}
for ``The Fox and the Grapes" development story, reproduced from Elson \shortcite{Elson12}, is shown in
Fig.~\ref{foxgrapes-sig}. The annotation process involves sequentially
labeling the original story sentences according to the {\sc sig}
formalism using Scheherazade's GUI. The annotators instantiate
characters and objects, assign actions and properties to them, and
provide their interpretation of {\it why} characters are motivated to
take the actions they do.  Scheherazade's GUI features a built-in
generation module in the what-you-see-is-what-you-mean paradigm ({\sc
  wysiwym}) to help na\"{i}ve users produce correct annotations by letting
them check the natural language realization of their encoding as they
annotate \cite{BouayadAghaetal00}). Scheherazade does have a built in 
surface generation engine, but it is inflexible with synonyms and syntax, thus why we are interested 
in utilizing {\sc personage}. As a baseline comparison for our method, we use Scheherazade's
generation engine to evaluate the correctness of {\sc personage}
outputs.  

\begin{wrapfigure}{l}{0.6\textwidth}
\begin{center}
\vspace{-0.4in}
\includegraphics[width=0.60\textwidth]{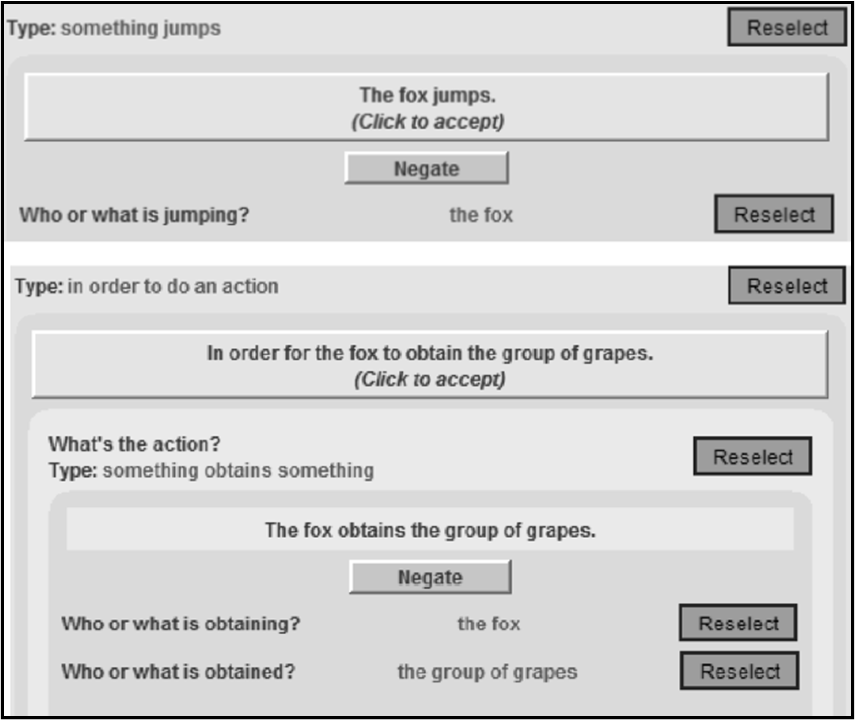}
\vspace{-0.3in}
\caption{\label{sig} GUI view of propositional modeling}
\vspace{-0.4in}
\end{center}
\end{wrapfigure}

One of the strengths of Scheherazade is that it allows users to
annotate a story along several dimensions, starting with the surface
form of the story (see first column in Fig.~\ref{foxgrapes-sig}) and
then proceeding to deeper representations.  The first 
dimension (see second column in Fig.~\ref{foxgrapes-sig})
 is called the ``timeline
  layer'', in which the story facts are encoded as predicate-argument
structures (propositions) and temporally ordered on a timeline.  The timeline layer consists of\\  \\a network of propositional structures, where nodes correspond to lexical items that are linked by
thematic relations. 
Scheherazade adapts information
about predicate-argument structures from the VerbNet lexical database \cite{Kipper06} and uses
WordNet \cite{WordNet} as its noun and adjectives taxonomy. The arcs
of the story graph are labeled with discourse relations.
Fig.~\ref{sig} shows a GUI screenshot of assigning propositional
structure to the sentence {\it The fox jumped in order to obtain the
  group of grapes}. This sentence is encoded as two nested
propositions {\tt jump(fox)} and {\tt obtain(fox, group of
  grapes)}. Both actions ({\tt jump} and {\tt obtain}) contain
references to the story characters and objects ({\tt fox} and {\tt
  grapes}) that fill in slots corresponding to semantic roles.

The second dimension (see third column in Fig.~\ref{foxgrapes-sig}) is
called the ``interpretative layer''; this layer goes beyond
summarizing the actions and events that occur, but attempts to capture
story meaning derived from agent-specific plans, goals, attempts,
outcomes and affectual impacts. 
To date, we only utilize the event timeline layer of the {\sc sig} encoding.

\begin{figure}[!htb]
\begin{scriptsize}
\begin{tabular}{|p{0.8in}p{0.6in}p{1.8in}p{1.5in}|c|c|c|c|}
\hline
{\bf Model} & {\bf Parameter} & {\bf Description} & {\bf Example}\\
\hline
\multirow{3}{*}{Shy voice} & {\sc Softener hedges} & Insert syntactic elements ({\it sort of},
{\it  kind of}, {\it somewhat}, {\it quite}, {\it around}, {\it
rather}, {\it I think that}, {\it it seems that}, {\it it seems to
me that}) to mitigate the strength of a proposition & {\it `It seems to me that he was hungry'}\\ 

&{\sc Stuttering} &  Duplicate parts of a content word & {\it `The vine hung on the tr-trellis'}\\

&{\sc Filled pauses} & Insert  syntactic elements expressing
hesitancy ({\it I  mean},  {\it err},  {\it mmhm}, {\it like}, {\it
you~know}) & {\it `Err... the fox jumped'}\\
\hline
\multirow{3}{*}{Laid-back voice} & {\sc Emphasizer hedges} & Insert  syntactic elements ({\it really},
{\it basically}, {\it actually}) to strengthen a
proposition & {\it `The fox failed to get the group of grapes, alright?'}  \\ 

& {\sc Exclamation} & Insert an exclamation mark & {\it `The group of grapes hung on the vine!'} \\

& {\sc Expletives} & Insert  a swear word & {\it `The fox was damn hungry'}\\
\hline
\end{tabular}
\end{scriptsize}
\vspace{-0.15in}
\centering \caption{\label{pers-fig} Examples of  pragmatic marker
insertion parameters from {\sc personage}}
\vspace{-0.2in}
\end{figure}

\noindent{\bf Personage.} {\sc personage} is an {\sc nlg} engine that
has the ability to generate a single utterance in many different
voices.  Models of narrative style are currently based on the
Big Five personality traits \cite{MairesseWalker11},
or are learned from film scripts \cite{Walkeretal11}.  Each type
of model (personality trait or film) specifies a set of language cues,
one of 67 different parameters, whose value varies with the
personality or style to be conveyed. Fig.~\ref{pers-fig} shows a
subset of parameters, which were used in stylistic models to produce ``The Fox and the Grapes'' in different voices (see Fig.~\ref{fox-grapes-retellings}).  Previous
work \cite{MairesseWalker11} has shown that humans perceive the personality stylistic models
in the way that {\sc personage} intended.

\begin{figure}[htb]
\begin{center}
\includegraphics[width=1.0\textwidth]{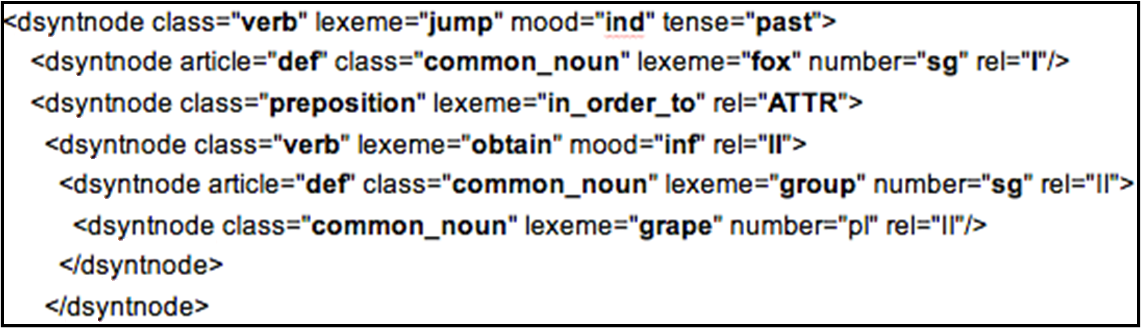}
\vspace{-0.2in}
\caption{\label{dsynts} DSyntS for a sentence {\it The fox jumped in order to obtain the group of grapes}}
\vspace{-0.3in}
\end{center}
\end{figure}

After selecting a stylistic model for each utterance, {\sc personage}
uses the off-the-shelf surface realizer {\sc RealPro}
\cite{LavoieRambow97}. {\sc personage} outputs a sentence-plan tree
whose internal representations are deep syntactic structures
(DsyntS) that RealPro expects as input.  DSyntS provides a flexible
dependency tree representation of an utterance which can be altered by
the {\sc personage} parameter settings.  
The nodes of the DSynts syntactic trees are labeled with lexemes and the
arcs of the tree are labeled with syntactic relations. 
The DSyntS formalism distinguishes between
arguments and modifiers and between different types of arguments
(subject, direct and indirect object etc). Lexicalized nodes also
contain a range of grammatical features used in generation.  RealPro
handles morphology, agreement and function words to produce an output
string.
Fig.~\ref{dsynts} shows an example DSyntS structure for one of the
sentences from our development story ``The Fox and the Grapes''. Feature values in bold need to be obtained from the internal Scheherazade
representation. 

\section{Method}
\label{method-sec}

\begin{figure}
\begin{centering}
\includegraphics[width = 0.8\textwidth]{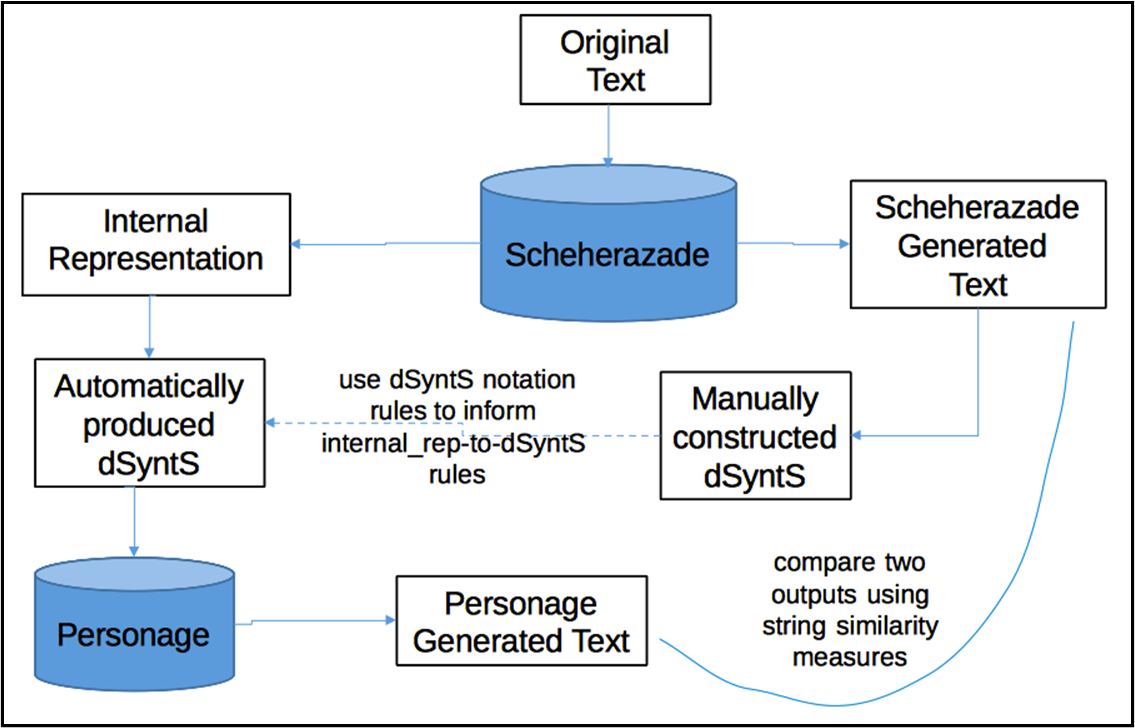}
\vspace{-0.1in}
\caption{\label{method} Our Method}
\vspace{-0.25in}
\end{centering}
\end{figure}

Our method applies a model of syntax to a
Scheherazade representation of a story (a {\sc sig} encoding), in order to
produce a retelling of the story in a different voice. 
A prerequisite for producing stylistic variations of a story is an ability to generate a ``correct'' retelling of the story. 
The focus of this study is to verify that the essence of a story is not distorted as we move from one formal representation of the story ({\sc sig}) to another (DSyntS). We use Scheherazade's built-in
generator, which was customized to the {\sc sig} schemata, as a baseline to evaluate our results.
Our data comes from DramaBank, a collection of Scheherazade annotated texts
ranging from Aesop fables to contemporary nonfiction
\cite{Elson12}. Aesop fables from DramaBank serve as our dataset: 
{\bf one} fable (seen in Fig.~\ref{fox-grapes-example}) is used in development, and then our method
is tested by automatically transforming the 35 other fables to the
{\sc personage} representation. Figs.~\ref{fox-grapes-example} and ~\ref{lion-boar-example} show two Aesop fables, with both Scheherazade and {\sc personage} generated versions. ``The Fox and the Grapes" was the
development story for our work, while ``The Lion and the Boar" was
part of the test set.
 Fig.~\ref{method} shows an overview of the method consisting of the following steps:

\begin{enumerate}
\item Use Scheherazade's built-in generation engine to produce text from the {\sc sig} encoding of the fable
\item Manually construct DSyntS corresponding to the text generated in step 1 (follow the right branch in Fig.~\ref{method}) 
\item Derive semantic representation of a fable from the {\sc sig} encoding using Schehe- razade API (follow the left branch of Fig.~\ref{method})
\item Informed by our understanding of the two formalisms, develop transformation rules to build DSyntS from semantic representation
\item Apply rules to the semantic representation derived in step 3
\item Feed automatically produced DSyntS to {\sc personage} using a neutral {\sc nlg} model and compare the output with that of step 1. The metrics we used for string comparison are discussed in Sec.~\ref{results-sec}.
\end{enumerate}

The primary technical challenge was developing a general mechanism for
converting {\sc sig} semantic encodings into the DSyntS representation
used by {\sc personage}'s generation dictionary.  This involved
enforcing syntactic tree structure over the chains of
propositions. The challenge was partly due to the variety of ways that
VerbNet and WordNet allow nuances of meaning to be expressed. For
example, a sentence {\it the crow was sitting on the branch of the
  tree} has two alternative encodings depending on what is important
about crow's initial disposition: the crow can {\it be sitting} as an
adjectival modifier, or can be {\it sitting} as a progressive
action. There are also gaps in Scheherazade's coverage of abstract
concepts, which can lead to workarounds on the part of the
annotators that are not easily undone by surface realization (an
example is {\it whether later first drank} in
Fig.~\ref{lion-boar-example}, where adjectival modifiers are used to
express {\it which of [the characters] should drink first} from the original text).

\begin{figure}
\centering
\includegraphics[width=1\textwidth]{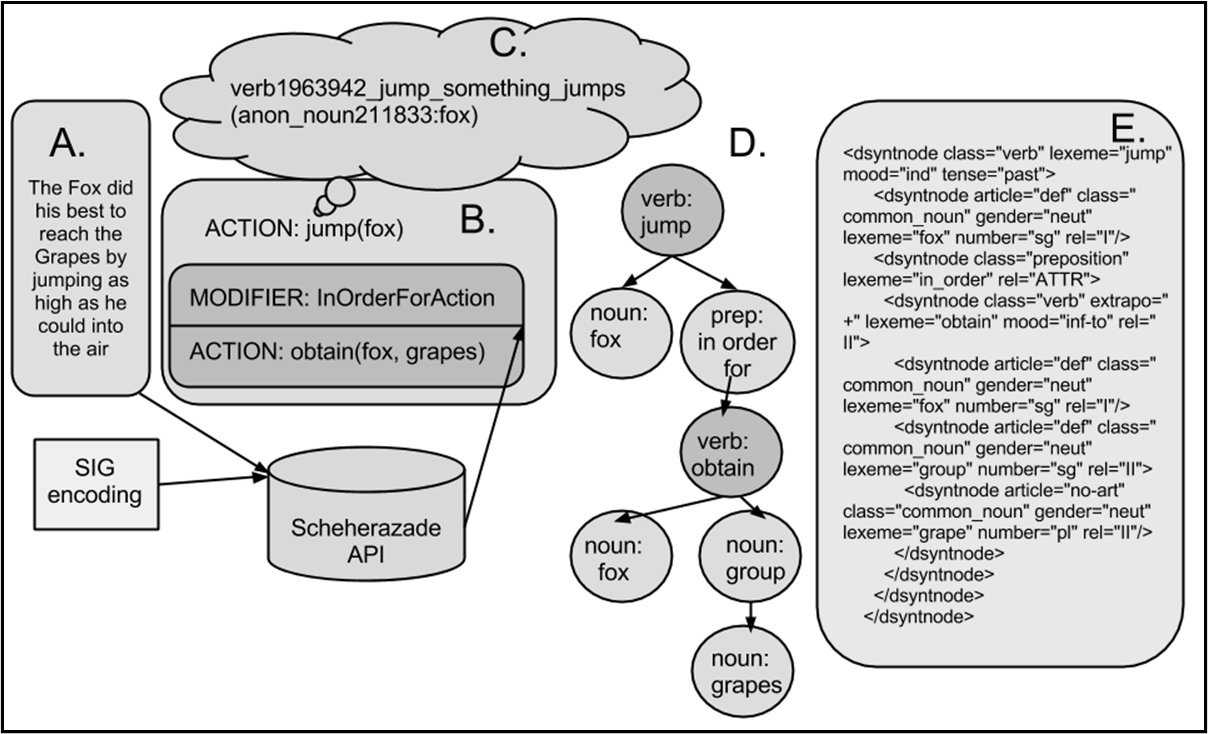}
\vspace{-0.15in}
\caption{\label{sig_dsynts} Step by step transformation from {\sc sig} to DSyntS}
\vspace{-0.2in}
\end{figure}

The transformation of Scheherazade's semantic representation into
syntactic dependency structure is a multi-stage process, illustrated
in Fig.~\ref{sig_dsynts}.  First, a syntactic tree is constructed
from the propositional event structure. Element A in
Fig.~\ref{sig_dsynts} contains a sentence from the original ``The
  Fox and the Grapes" fable.  We use the Scheherazade API to process
the fable text together with its {\sc sig} encoding and extract
actions associated with each timespan of the timeline layer.  Element
B in Fig.~\ref{sig_dsynts} shows a schematic representation of the
propositional structures.  Each action instantiates a separate tree
construction procedure.  For each action, we create a verb instance
(see highlighted nodes of element D in Fig.~\ref{sig_dsynts}). We
use information about the predicate-argument frame that the action invokes (see
element C in Fig.~\ref{sig_dsynts}) to map frame constituents into
respective lexico-syntactic classes, for example, characters and
objects are mapped into nouns, properties into adjectives and so
on. The lexico-syntactic class aggregates all of the information that
is necessary for generation of a lexico-syntactic unit in DSyntS 
(element E in Fig.~\ref{sig_dsynts}). We
define 6 classes corresponding to main parts of speech: noun, verb,
adverb, adjective, functional word. Each class has a list of
properties such as morphology or relation type that are required by
the DSyntS notation for a correct rendering of a category. For
example, all classes include a method that parses frame type in the {\sc
  sig} to derive the base lexeme. The methods to derive grammatical
features are class-specific. Each lexico-syntactic unit refers to the
elements that it governs syntactically thus forming a hierarchical
structure. A separate method collects the frame adjuncts as they have
a different internal representation in the {\sc sig}.

At the second stage, the algorithm traverses the syntactic tree in-order
and creates an XML node for each lexico-syntactic unit. Class
properties are then written to disk, and the resulting file (see element E in Fig.~\ref{sig_dsynts}) is
processed by the surface realizer to generate text.
Fig.~\ref{lion-boar-example} shows the ``Lion and the Boar"
fable from our test set, with its generated versions.


\begin{figure}[t!hb] 
\begin{center}
\begin{scriptsize}
\begin{tabular}{|p{1.3in}|p{1.8in}|p{1.3in}|}
\hline
{\bfseries Original Story}  & {\bfseries Scheherazade} & {\bf Personage} \\ \hline \hline
On a summer day, when the great heat induced a general thirst, a Lion
and a Boar came at the same moment to a small well to drink. 
They fiercely disputed which of them should drink first, and were soon
engaged in the agonies of a mortal combat. 
On their stopping on a
sudden to take breath for the fiercer renewal of the strife, they saw
some Vultures waiting in the distance to feast on the one which should
fall first. 
They at once made up their quarrel, saying: ``It is better
for us to make friends, than to become the food of Crows or Vultures,
as will certainly happen if we are disabled.''

& Once, the air was hot.  
A boar decided to drink from a spring, and a lion decided to drink from the spring.  
The boar quarrelled about whether later first drank from the spring, and the lion quarrelled about whether later first drank from the spring.  
The boar began to attack the lion, and the lion began to attack the boar.  
The boar stopped attacking the lion, and the lion stopped attacking the boar.  
The boar above saw a group of vultures being seated on some rock, and the lion above saw the group of vultures being seated on the rock.  
The group of vultures began to plan -- if the boar were to later die, and the lion were to later die -- for the group of vultures to eat.  
The boar sobered, and the lion sobered.  The boar said to the lion
that -- if the lion were to not kill the boar -- the group of vultures
would not eat the boar, and the lion said to the boar that -- if the
boar were to not kill the lion -- the group of vultures would not eat
the lion.
The boar didn't kill the lion, and the lion didn't kill the boar.

& The air was hot.
       The lion decided the lion drank from the spring.
       The boar decided the boar drank from the spring.
       The boar quarreled.
       The lion quarreled.
       The lion attacked the boar.
       The boar attacked the lion.
       The boar above saw the group of vultures was seated on the rock.
       The lion above saw the group of vultures was seated on the rock.
       The group of vultures planned the group of vultures ate.
       The boar sobered.
       The lion sobered.
       The boar said the group of vultures did not eat the boar to the lion.
       The lion said the group of vultures did not eat the lion to the boar.
       The boar did not kill the lion.
       The lion did not kill the boar.
\\ \hline
\end{tabular}
\end{scriptsize}
\vspace{-0.1in}
\caption{\label{lion-boar-example} 
``The Lion and the Boar" Fable from the test set, with Generated Versions}
\vspace{-0.1in}
\end{center}
\end{figure}


\begin{figure}[t!hb] 
\begin{center}
\begin{scriptsize}
\begin{tabular}{|p{1.5in}|p{1.6in}|p{1.5in}|}
\hline
{\bfseries Personage FORMAL}  & {\bfseries Personage SHY} & {\bf Personage LAID-BACK} \\ \hline \hline

The group of grapes hung on the vine. The vine hung on the trellis.The fox saw the group of grapes. The fox jumped in order to obtain the group of grapes.The fox didn't obtain the group of grapes because he couldn't reach the group of grapes.
& 

Well, the group of grapes hung on the vine. The vine hung on the tr-tr-trellis. The fox saw the group of grapes. It seemed that he was hungry. Err... the fox jumped in order to obtain the group of grapes. The fox didn't collect the group of grapes.

&

Ok, the group of grapes rested on the vine. The vine hung on the trellis, didn't it? The fox saw the group of grapes. He damn was hungry. The fox jumped in order to obtain the group of grapes, okay? Oh the fox failed to get the group of grapes, didn't obtain it, because he couldn't reach the group of grapes, you see?

\\ \hline
\end{tabular}
\end{scriptsize}
\vspace{-0.1in}
\caption{\label{fox-grapes-retellings} 
Retellings of ``The Fox and the Grapes"}
\vspace{-0.3in}
\end{center}
\end{figure}

Although the emphasis of this work is on generating a ``correct'' retelling of a story, this infrastructure allows us to alter the story stylistically.
Fig.~\ref{fox-grapes-retellings} illustrates how we can now piggy-back
on the transformations that {\sc personage} can make to produce
different retellings of the same story. The {\sc formal} voice triggered no stylistic parameters of {\sc personage}. ``The Fox and the Grapes'' was generated directly from DSyntS by the surface realizer. Fig.~\ref{pers-fig} provides examples of how different parameters played out for the {\sc shy} and {\sc laid-back} voices. The corresponding renderings of the story demonstrate lexical ({\it grapes hung}/{\it grapes rested}), syntactic ({\it didn't get}/{\it failed to get}) and stylistic variation.


\section{Results}
\label{results-sec}


To evaluate the perfomance of our translation rules
we compare the output generated by {\sc personage} to that of Scheherazade's built-in realizer, using two metrics: BLEU score \cite{Papineni02} and Levenshtein distance. {\it BLEU} is an established standard for evaluating the quality of machine translation and summarization systems. The score between 0 and 1 measures the closeness of two documents by comparing n-grams, taking word order into account. {\it Levenshtein distance} is the minimum edit distance between two strings. The
objective is to minimize the total cost of character deletion,
insertion, replacement that it takes to transform one string into
another. In our case, we treat each word as a unit and measure word
deletion, insertion, and replacement. We used word stemming as a preprocessing step
to reduce the effect of individual word form variations. The results are shown in
Table~\ref{statistics}. {\bf Scheherazade-Personage} compares the output
of the {\sc personage} generator produced through our automatic
translation rules to that of the Scheherazade generation. Because the
rules were created on the development set to match the Scheherazade
story, we expect these results to provide a topline for comparison to
the test set, shown in the bottom table of Table~\ref{statistics}.

\begin{table}
\begin{center}
\caption{Mean and Standard Deviation for Levenshtein Distance (Lower is Better), and  
BLEU (Higher is Better) on both the DEVELOPMENT and TEST sets\label{statistics} }
\begin{tabular}{|l|c|c|}
\hline
DEVELOPMENT & Levenshtein & BLEU \\ \hline \hline
Scheherazade-Personage 		& 31 & .59 \\ \hline
Fable-Scheherazade		& 80 & .04 \\ \hline
Fable-Personage		& 84 & .03 \\ \hline
\multicolumn{3}{c}{} \\ 
\multicolumn{3}{c}{} \\ \hline
TEST & Levenshtein Mean (STD) & BLEU Mean (STD) \\ \hline \hline
Scheherazade-Personage 		& 72 (40) & .32 (.11) \\ \hline
Fable-Scheherazade		& 116 (41) & .06 (.02) \\ \hline
Fable-Personage		& 108 (31) & .03 (.02)\\ \hline
\end{tabular}
\vspace{-0.3in}
\end{center}
\end{table}



Although our rules were informed by looking at the Scheherazade
generation, Table~\ref{statistics} also includes measures of the
distance between the original fable and both Scheherazade and {\sc
  personage}. {\bf Fable-Scheherazade} and {\bf Fable-Personage} compare
the original fable to the Scheherazade and to the {\sc personage} generation
respectfully. Note that these results should not be compared to {\bf
  Scheherazade-Personage} since both of these are outputs of an {\sc
  nlg} engine. The two-tailed Student's t-test was used to compare the two realizers' mean distance values to the original fables and determine statistical significance. 
The difference in Levenshtein distance between {\bf Fable-Scheherazade} and {\bf Fable-Personage} 
 is not statistically significant (p = 0.08) on both development and test sets.
This indicates that our rules generate a story which is
similar to what Scheherazade generates in terms of closeness to the
original. However, Scheherazade shows a higher BLUE score with the original fables (p $<$ 0.001). We believe that this is due to the fact that our translation rules assume a simple generation. 
The rules did not attempt to express tense, aspect or stative/non-stative
complications, all of which contribute to a lower overlap of
n-grams. The assignment of tense and aspect was an area of particular
focus for Scheherazade's built-in realizer \cite{ElsonMcKeownTense}.

\section{Conclusions and Future Work}
\label{conc-sec}
In this paper we show that: (1) Scheherazade's {\sc sig} annotation schemata provides a
rich and robust story representation that can be linked to other
generation engines; (2) we can integrate Scheherazade and {\sc
  personage} (two off-the-shelf tools) for representing and producing
narrative, thus bridging the gap between content and sentence planning
in the {\sc nlg} pipeline; and (3) we have created an infrastructure which
puts us in a position to reproduce a story in different voices and
styles.  In this paper, we presented quantitative results using Levenshtein
distance and BLEU score. However, since our long term goal is to
generate different retellings, these metrics will be
inappropriate. Here we were primarily concerned with the correctness
of the content of the generators; in future work we will need to develop
new metrics or new ways of measuring the quality of story
retellings. In particular we plan to compare human subjective
judgements of story quality across different retellings.



There are also limitations of this work.  First, the {\sc personage}
realizer needs to extend its generation dictionary in order to deal
with irregular forms. The current system generated incorrect forms
such as {\it openned}, and {\it wifes}.  Also we were not able to make
tense distinctions in our generated version, and generated everything
in past simple tense. In addition, we noted problems with the
generation of distinct articles when needed such as {\it a} vs. {\it
  the}.  There are a special set of surface realization rules in
Scheherazade that are currently missing from {\sc personage} that adds
cue phrases such as {\it that} and {\it once}.  We aim to address
these problems in future work. 

It should be mentioned that despite being domain independent, our
method relies on manual story annotations to provide content for the
generation engine. DramaBank is the result of a collection experiment
using trained annotators; as they became familiar with the tool, the
time that the annotators took to encode each fable (80 to 175 words)
as a {\sc sig} encoding dropped from several hours to 30-45 minutes on
average \cite{Elson12}. The notion of achieving the same semantic
encoding using automatic methods is still aspirational. While the {\sc
  sig} model overlaps with several lines of work in automatic semantic
parsing, as it has aspects of annotating attribution and private
states \cite{Wiebe}, annotating time \cite{Pustejovsky03} and annotating verb frames and semantic roles \cite{PropBank}, there is not yet a semantic parser that can
combine these aspects into an integrated encoding, and developing one
falls outside of scope of this work.

In future work, we also aim to do much more detailed studies on the
process of generating the same story in different voices, using the
apparatus we present here. Examples of stylistic story variations
presented in this paper come from modifications of narrator's
voice. In future work, we plan to apply stylistic models to story
characters. An extension to Scheherazade to distinguish direct and
indirect speech in the {\sc sig} will allow give characters
expressive, personality driven voices. Once a story has been modeled
symbolically, we can realize it in multiple ways, either by different
realizers or by the same realizer in different modes.



\bibliographystyle{splncs}


\end{document}